\documentclass[letterpaper, 10pt, conference]{ieeeconf}  

\IEEEoverridecommandlockouts                              

\usepackage{graphics} 
\usepackage{amsmath} 
\usepackage{amsfonts}
\usepackage{csquotes}
\usepackage{textcomp, gensymb} 
\usepackage{subcaption}
\usepackage{array}
\newcolumntype{P}[1]{>{\centering\arraybackslash}p{#1}}  
\usepackage{booktabs}
\usepackage{graphicx}
\usepackage{mathtools}
\usepackage{copyright} 

\title{\LARGE \bf
CAR-DESPOT: Causally-Informed Online POMDP Planning\\
for Robots in Confounded Environments
}

\author{Ricardo Cannizzaro and Lars Kunze
\thanks{This work is supported by the Australian Defence Science \& Technology Group and the EPSRC RAILS project (grant reference: EP/W011344/1).}
\thanks{The authors are with the Oxford Robotics Institute, Dept. Engineering Science, University of Oxford, UK. {\tt\small ricardo@robots.ox.ac.uk}}%
}

\begin{document}

\maketitle
\copyrightnotice   

\begin{abstract}
Robots operating in real-world environments must reason about possible outcomes of stochastic actions and make decisions based on partial observations of the true world state.
A major challenge for making accurate and robust action predictions is the problem of confounding, which if left untreated can lead to prediction errors.
The partially observable Markov decision process (POMDP) is a widely-used framework to model these stochastic and partially-observable decision-making problems. However, due to a lack of explicit causal semantics, POMDP planning methods are prone to confounding bias and thus in the presence of unobserved confounders may produce underperforming policies.
This paper presents a novel causally-informed extension of \enquote{anytime regularized determinized sparse partially observable tree} (AR-DESPOT), a modern anytime online POMDP planner, using causal modelling and inference to eliminate errors caused by unmeasured confounder variables.
We further propose a method to learn offline the partial parameterisation of the causal model for planning, from ground truth model data.
We evaluate our methods on a toy problem with an unobserved confounder and show that the learned causal model is highly accurate, while our planning method is more robust to confounding and produces overall higher performing policies than AR-DESPOT.
\end{abstract}


\section{INTRODUCTION}



Robots must be capable of making decisions to achieve their tasks efficiently and safely, especially in safety-critical and high-risk applications. 
This must be done in a manner that is robust to the range of sources of uncertainty typically encountered in real-world environments with stochastic action outcomes, and partial and noisy sensor observations \cite{Kurniawati2022POMDPReview}.

The partially observable Markov decision process (POMDP) is a popular framework that provides a principled approach for formulating sequential decision-making problems for agents operating in stochastic and partially-observable environments \cite{Smallwood1973}. 
It has been successfully demonstrated in a wide range of robotics domains \cite{Bai2015DESPOTAutonomousDriving}\cite{Budd2022}\cite{Goldhorn2018SearchAndTrack}.

Despite this, standard POMDPs lack the causal semantics required to articulate system relationships in a causal manner and thus the application of causal analysis is not possible. 
This is problematic when unobserved confounding between the agent's action selection and outcome is present in the system. Confounding occurs when there is a common cause of the target cause and effect variables, which introduces a spurious correlation between variables that is entangled with the true causation we are interested in \cite{Pearl2009Causality}. Without the proper causal treatment, this can lead to prediction errors caused by confounding bias. 
For POMDP models where the confounder is unobservable (UCPOMDPs), it is not possible to control for it in analysis. In these cases standard POMDP planners may generate underperforming policies \cite{Bareinboim2015ConfoundedBandits}\cite{Zhang2016CausalMDPUC}.

This has significant implications for robots operating in real-world environments. Sources of confounding can arise from the environment, which system designers may not have anticipated and/or are not observable to the robot. Examples include weather phenomena (e.g., wind), surface conditions (e.g., slippery surfaces), extraneous factors (e.g., magnetic fields, radiation sources), and adversarial agents whom may manipulate conditions to hinder the robot's progression. 

\begin{figure}
    \centering
    \begin{subfigure}[b]{0.40\columnwidth}
        \centering
        \includegraphics[height=4.0cm]{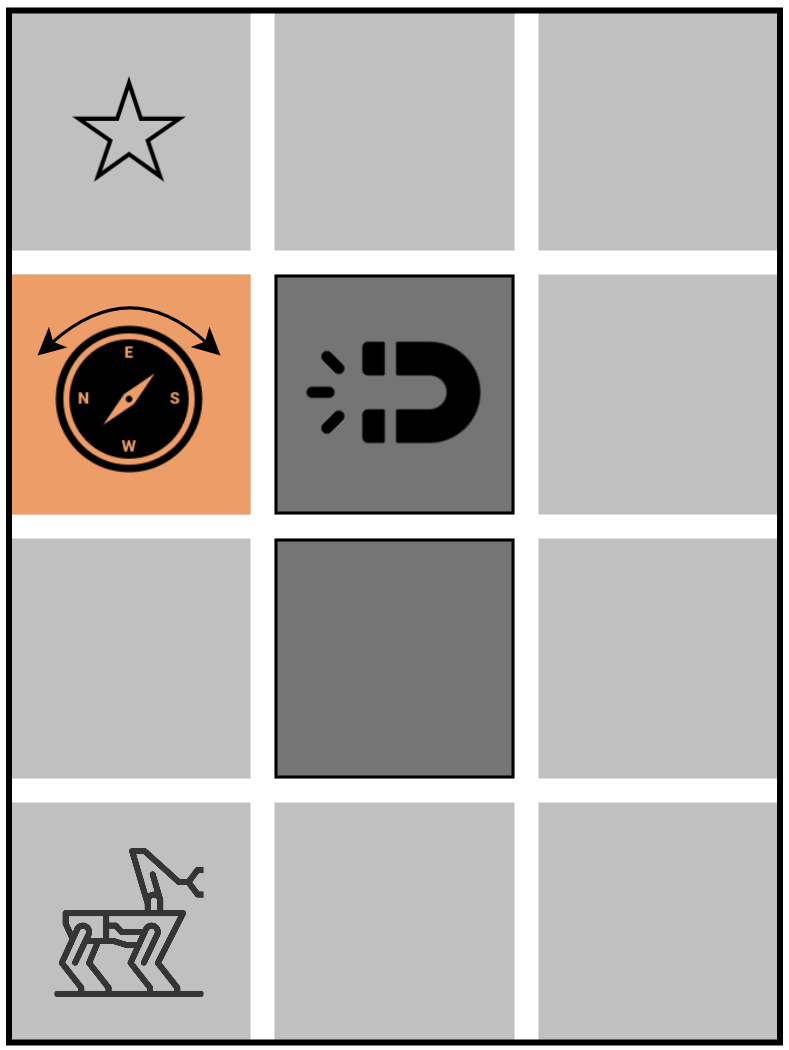} 
	    \caption{}
         \label{fig:grid_world_confounded}
    \end{subfigure}
    \hfill
    \begin{subfigure}[b]{0.52\columnwidth}
         \centering
         \includegraphics[height=4.0cm]{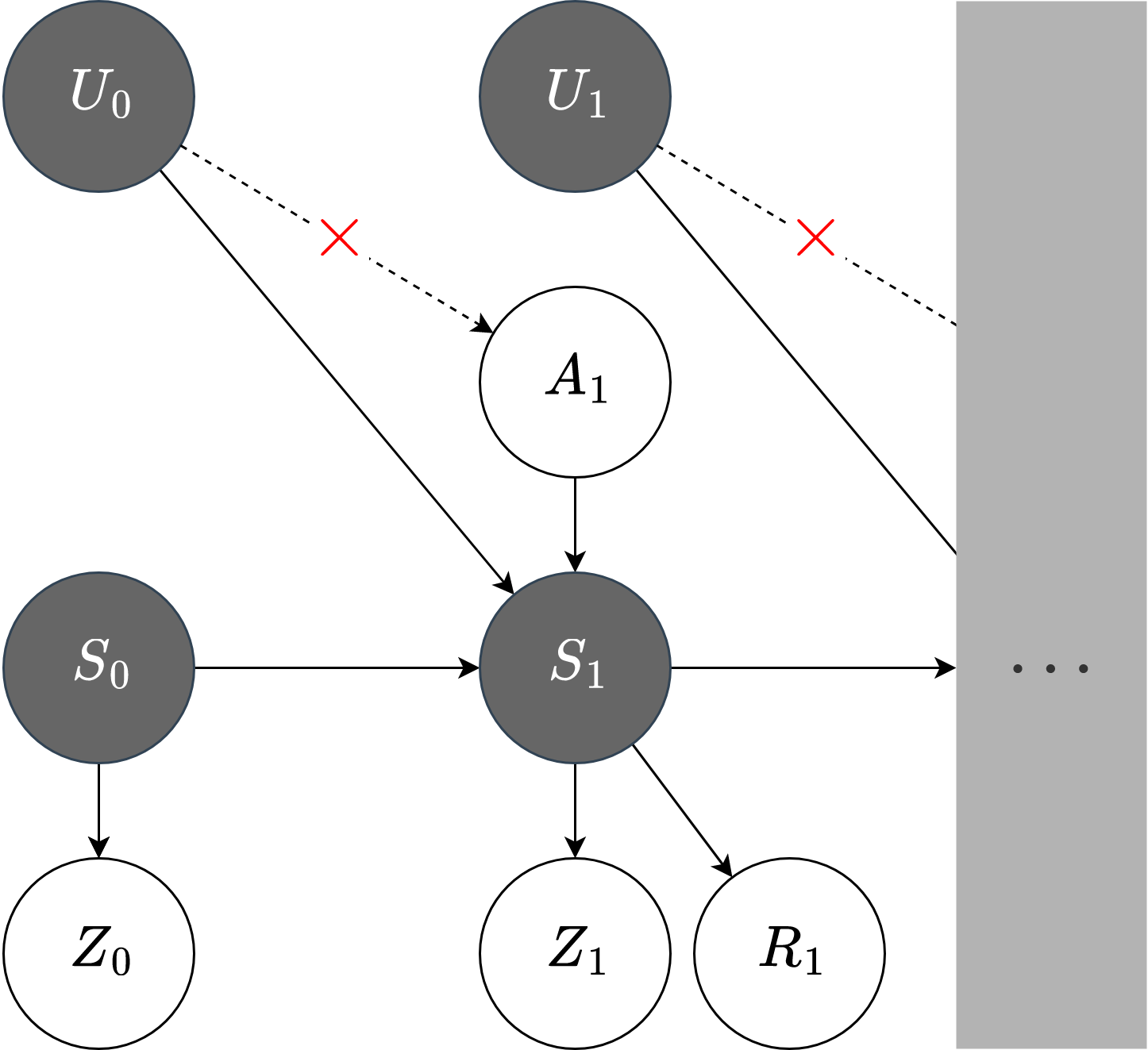}
        \caption{}
		\label{fig:scm_ucpomdp_intervened}
    \end{subfigure}
    \hspace{0.02\columnwidth}
    \caption[GridWorldConfounded Problem.]{(a) \emph{GridWorldConfounded}. The robot moves in the four cardinal directions in a grid to reach the goal while avoiding occupied cells. Variable sensor interference (orange cell) from an electromagnet is an unobserved confounder. 
    (b) The SCM representation of a UCPOMDP. The agent's reactive action selection and state updates are both conditionally dependent on the unobserved confounder. Causal interventions (red crosses) are taken on actions to 
    eliminate confounding bias in transition predictions. Observed variables are shown as white nodes, unobserved as grey.
    }
    \label{fig:pg_1_fig}
    \vspace{-15pt}  
\end{figure}

To address this problem, this paper proposes a novel causally-informed extension of the efficient anytime regularized determinized sparse partially observable tree (AR-DESPOT) online planner \cite{Ye2017DESPOT}\cite{Somani2013DESPOT}, that uses causal modelling and inference to eliminate policy bias errors caused by unmeasured confounding variables. To address the further problem that the system dynamics model may not be fully known for planning \textit{a priori}, we also propose a method to learn offline the partial model parameterisation from data.

We experimentally evaluate our learning and planning methods on a toy problem with an unobserved confounder to demonstrate the benefits of combining casual modelling and inference with POMDP planning. While we are motivated by real-world robotics problems, we start here with a minimal toy problem representative of an autonomous inspection mission as an initial proof of concept.

The main contributions of this work are: 1) the novel formulation of the CAR-DESPOT algorithm; 2) a quantitative evaluation on a UCPOMDP toy problems; and 3) a demonstration of how components of the causal model can be learned offline from ground truth observational data. 

\section{RELATED WORK}\label{sec:RelatedWork}

\subsection{Causality in Robotics}
Recent work has identified a growing need for causality as a component of robotics and trustworthy autonomous systems \cite{Ganguly2023}\cite{Hellstrom2021CausationInRobotics}\cite{Lake2017BuildingMachines}. While research into causality in robotics is still nascent, it has been proposed to hold great potential benefits such as addressing machine learning domain transfer issues by providing artificial agents an understanding of cause and effect \cite{Pearl2019CausalInferenceAndML}. Further, they can be used to achieve desirable properties of autonomous systems, including explainability, fairness, and accountability \cite{Ganguly2023}, which is typically not possible with \enquote{black-box} deep learning methods. 

Existing works on causal discovery in robotics aim to discover action-outcome causal relationships, including from human demonstrations in table-setting tasks \cite{Uhde2020RobotAsScientist} and from simulated robot actions in block stacking tasks \cite{Diehl2022WhyDidIFail}. In the former, authors learn temporal co-occurrences to approximate causal relationships, while in the latter authors learn the structure and parameterisation of a causal Bayesian network. In our work, we address a sub-problem of causal discovery: causal parameterisation learning. Further, we represent the system as a structural causal model (SCM) \cite{Pearl2009Causality}, which is a more expressive type of causal model that also permits the more sophisticated counterfactual inference (see \ref{subsec:causality}). 

 Causal reasoning has been applied to planning problems in earlier works. One approach uses causal action models to create plans for office delivery jobs \cite{Beetz1998Causal}, while another uses causal reasoning for coordination of multiple household cleaning robots \cite{Aker2011Causal}. More recent work combines causal learning and reasoning to explain, predict, and prevent plan failures in robotic block stacking tasks \cite{Diehl2023ExplainPredictPreventFailures}. 
 None of these approaches adopt a principled POMDP/MDP formulation used in our work, nor do they incorporate any probabilistic causal machinery \cite{Pearl2009Causality} as we do. To the best of our knowledge, we are not aware of any existing online POMDP-based robot planner that utilises causal modelling and inference. 

\subsection{POMDP-Based Robot Planning}
 In POMDP planning problems, the agent must find an optimal (or near-optimal) policy, which maps its belief over world states to the action to take to optimise its long-term reward. The true world state is unknown to the agent. Instead, it must rely on partial observations of the world to update its belief, which is a probability distribution over the possible states, and reason over this belief. POMDPs are extensively used in the probabilistic planning literature; a review of POMDPs in robotics is presented in \cite{Kurniawati2022POMDPReview}.

Online policy search is a POMDP planning method that scales well to larger POMDPs \cite{Kurniawati2022POMDPReview}. Online planning alternates between policy planning and execution, with the agent searching for only the immediate next-best action at each time step. 
%
Several online planners have achieved notable success in robot POMDP planning problems, including POMDP-lite \cite{Chen2016POMDPLite}, Adaptive Belief Tree \cite{Kurniawati2016}, POMCP \cite{Silver2010POMCP}, and AR-DESPOT \cite{Somani2013DESPOT}. However, none of these POMDP planners use explicitly causal models, thus when used for planning in environments with unobserved confounders, they are prone to confounding bias and thus may generate sub-optimally performing policies \cite{Zhang2016CausalMDPUC}.

\subsection{Causal Reinforcement Learning}
Recent work has begun bridging the gap between reinforcement learning (RL) and causal inference research communities under the umbrella of causal RL \cite{Bareinboim2015ConfoundedBandits}\cite{Zhang2016CausalMDPUC}\cite{Ortega2021}\cite{Gasse2021}.

While we adopt a SCM-based (PO)MDP formulation alike these works, they differ from ours in that they address model-based reinforcement learning problems, which involve two sub-problems: 1) learning the transition and reward distributions; and, 2) deciding on the optimal action sequence (i.e., planning). Thus, online RL-based methods inherently require the agent to take experiments in the world to learn beneficial state-observation sequences, which is not always feasible in mobile robotics applications due to safety constraints (e.g., autonomous driving), where robot recovery is not practical (e.g., underwater or planetary robots), or where the damage to the robot and infrastructure is too costly to allow the robot to fail while learning the value of poor action-observation sequences (e.g., industrial inspection). Thus, we focus on the application of causality to planning, in mobile robotics applications for which online experimentation is not suitable.

\section{BACKGROUND KNOWLEDGE}
\subsection{Causal Modelling and Inference}\label{subsec:causality}
Causal models encode the causal relationships that describe the data-generation process of a system. Critically, these are causal rather than merely association-based. As such, they encode relationships where the flow of information is strictly in the direction from cause to effect.

Judea Pearl's \emph{Ladder of Causality} defines a hierarchy of three causal inference types of increasing complexity \cite{Pearl2018Book}. The first level, \emph{association}, is the most primitive and relates to the activity of seeing. Queries of this type involve reasoning about how taking an observation of one variable ($X$) should change our belief in another related variable ($Y$); e.g., $P(Y|X)$. The second level, \emph{intervention}, relates to the activity of doing; that is taking an action to externally intervene on the system and interrupt the natural causal relationships that govern the data-generation process. 
This is represented with the $do(\cdot{})$ operator \cite{Pearl2009Causality}, which defines an intervention to modify a set of variable outcomes in the causal model. Interventional inference queries involve reasoning about how taking an action to set a variable ($X$) to a fixed value ($x$) should change our belief in another; e.g., $P\big(Y|do(X=x)\big)$. The third level, \emph{counterfactual}, relates to the activity of imagining other worlds with outcomes other than what we have observed (i.e., worlds that are counter to our factual evidence). 

\textbf{Causal Bayesian networks (CBNs)} and \textbf{structural causal models (SCMs)} are two common generative causal probabilistic graphical model types used in causal modelling and causal inference \cite{Pearl2009Causality}. For both, causal relationships are represented by an underlying causal directed acyclic graph (DAG). A causal DAG $G$ is defined as $G=\langle V, E \rangle$, where $V$ is the set of nodes, representing random variables in the causal model, and $E$ is the set of directed edges, the presence of which between two variables denoting the existence of a conditional dependency of the child on the parent node. 

In the case of CBNs, a causal DAG is combined with conditional probability distributions. 
SCMs are also constructed from an underlying casual DAG $G$, but are different to CBNs in that they separate the stochastic elements of a model from the deterministic ones \cite{Pearl2009Causality}. This permits their use in answering counterfactual queries, which cannot be done with CBNs alone. A structural causal model $M$ is defined as a 4-tuple $M=\langle U, V, F, P(U)\rangle$, where $U$ is a set of exogenous random variables that are determined by factors external to the model, $V$ is a set of endogenous variables, $F$ is a set of deterministic functions used to assign the endogenous variable $v_i$ its value as a function of its graph parents $pa_i$ and corresponding exogenous variable $u_i$,
and $P(U)$ is a probability distribution over the exogenous random variables:
\begin{equation}
    u_{i} \sim P(U_i)
    \quad \quad
    v_i := f_i(u_i,{pa}_i)
    \label{eq:scm_equations}
\end{equation}

Through conditioning and intervention operations, CBNs and SCMs can be used to answer observational and interventional inference queries, which in general cannot be expected to coincide.
For models with confounding variables, the \textbf{intervention} operation disentangles the spurious correlation from the true causation by blocking the influence of the confounding parent variables in the graph. Conversely, the \textbf{observational} distribution contains both the true causation and spurious correlation introduced by the confounder and thus it may provide a biased estimate of causal effects. 


\subsection{POMDP Planning}
A POMDP is defined as a tuple $\langle S, A, T, Z, O, R \rangle$, where $S$ is a set of world states, $A$ is a set of actions available to the agent, 
$T$ is the state-transition function $\big(T(s,a,s^\prime)\big)$ specifying the probability distribution over arriving in successor state $s^\prime$ after taking action $a$ from current state $s$, $Z$ is a set of observations the agent can take,
$O$ is the observation function specifying the probability distribution $\big(O(s^\prime,a,z)\big)$ over possible observations $z$ made after taking a given action $a$ and arriving in a given state $s^\prime$, and 
$R$ is the reward function specifying the immediate reward received by taking an action in a given state and arriving in a successor state. 

During policy search and execution, the agent starts with an initial belief $b_0$. After taking an action at $t-1$ the agent receives an observation $z_t$ and updates its belief $b_t$ using Bayes' Rule. A policy $\pi : B \rightarrow A$ defines the action $a \in A$ to take when the agent holds the belief $b \in B$. The value $V$ of a policy $\pi$ at a belief $b$ can be defined as the expected total discounted reward for following $\pi$ given the initial belief $b$: $V(b)=E\bigr[ \sum_{t=0}^{\infty} \gamma^t R\big(s_t, \pi(b_t)\big) | b_0=b \bigr]$, given a discount factor $0 < \gamma < 1$ applied to reduce the value of rewards further into the future. 
Using the \textit{argmax} decision rule, the optimal \textit{greedy} policy $\pi^*$ is the policy $\pi$ for which the policy value $V_\pi$ is maximised for a given initial belief $b_0$.
	

Constructing a belief tree is one method for online planning. By rooting the tree at the current belief $b_0$ and performing a lookahead search, a policy $\pi$ can be found that maximises $V_\pi(b_0)$. 
With each node in the tree representing a belief, a node branches into an edge for each action and branches again for each observation. Trials are conducted by simulating different possible sequences of alternating actions and observations.
The policy with the highest expected value can be found by applying the Bellman optimality principle to recursively calculate the maximum value of action branches and average value of observation branches.
Completing the belief tree search produces an approximately optimal policy.

\subsection{AR-DESPOT Planner}\label{sec:AR-DESPOT}
AR-DESPOT \cite{Ye2017DESPOT}\cite{Somani2013DESPOT} is an anytime online POMDP algorithm that addresses key computational challenges with planning under uncertainty, which can in the worst case render POMDP planning computationally intractable \cite{Madani1999Undecidability}
While a standard belief tree includes executions for all possible combinations of actions and observations, a DESPOT contains only a set of randomly sampled \textit{scenarios}. Thus, a search through a DESPOT is more efficient, as it focuses only on a subset of the complete belief tree. By searching through a DESPOT instead of a standard belief tree, AR-DESPOT is able to solve difficult POMDP problems with extremely large observation or state spaces \cite{Somani2013DESPOT}.

Since constructing a full DESPOT is not practical for very large POMDPs, the AR-DESPOT algorithm performs anytime heuristic search. By constructing the DESPOT incrementally following a heuristic, AR-DESPOT is able to scale up to very large POMDPs by focusing search on promising areas of the DESPOT. This balance of optimality and efficiency motivates its use in our work.

\section{PROPOSED METHOD FOR MODEL LEARNING AND CAUSAL PLANNING}
Our proposed method consists of two components: 1) learning the parameterisation of the SCM model transition function; and, 2) using the learned model and interventional inference to improve POMDP planning performance in the presence of an unobserved confounder. We begin by describing our adopted causal representation of the POMDP model.

\subsection{Causal POMDP Representation}\label{subsec:CausalPOMDPRepresentation}

%
We adopt a SCM-based formulation of POMDPs to permit interventional and counterfactual inference similar to the SCM-based MDP formulation in \cite{Zhang2016CausalMDPUC}. Although in this work we focus on interventional inference, this formulation sets up future work to apply counterfactual inference methods.

An \textbf{SCM-POMDP} is an SCM augmented with POMDP components, such that the set of endogenous variables $V$ contains the world state $s_k \in S$ at time $k$; the agent's choice of action $a_{k+1} \in A$; and successor state $s_{k+1}$, observation generated $z_{k+1} \in Z$, and immediate reward $r_{k+1}$ received after taking the action. The system's transition and observation probability functions and reward assignment are split into stochastic and deterministic components, respectively encoded by the set of exogenous variables $P(U)$ and assignment functions $F$. 

POMDPs with unobserved confounding (\textbf{UCPOMDPs}) are an extension of POMDPs, in which there is a set of unobserved confounding variables $U_{k}$ that influence both the agent's choice of action $A_{k+1}$ and the successor state $S_{k+1}$. \textbf{SCM-UCPOMDPs} are UCPOMDPs represented as SCMs. This is illustrated in Fig. \ref{fig:scm_ucpomdp_intervened}.
Critically, the agent's choice of action is not an independent distribution, as it is in the SCM-POMDP formulation. Instead, it is conditionally dependent on one or more confounding variables.
This dependence of the agent's action selection on unobserved confounder variables represents the \emph{reactive} tendencies of the agent; that is, how the agent is likely to act reflexively in response to the external influences. 
This is in contrast to the \emph{deliberative} decision-making process of the agent, in which actions are made independent of external influences. The exogenous variable distributions $P(U)$ and assignment functions $F$ for a SCM-UCPOMDP (Eq \ref{eq:scm_equations}) follow the structure of the SCM-UCPOMDP causal DAG shown in Fig \ref{fig:scm_ucpomdp_intervened}.

We represent the transition function under the agent's reactive behaviour using the observational distribution $P(S^\prime|A,S)$. Here, the back-door path introduced by the confounder means that the observational distribution contains the spurious correlation through $A\leftarrow U \rightarrow S^\prime$ in addition to causal influence directly through $A\rightarrow S^\prime$. In the case of deliberative decision-making, we use causal semantics to represent the transition function using the interventional distribution $P(S^\prime|do(A=a),S)$. Here, the $do$-operator is used to surgically remove the influence of the confounding variables on the agent's choice of action, thus retaining only the causal influence $A\rightarrow S^\prime$.

\subsection{Model Parameter Learning Method}
Our proposed method aims to learn the parameterisation of an SCM-UCPOMDP model from a ground truth model in a privileged setting. We are motivated by real-world robotics applications where the full system dynamics model is not fully known \textit{a priori}, but instead needs to created using a hybrid method combining domain knowledge and observational data in an offline setting prior to robot deployment. 

We define a ground truth SCM-UCPOMDP model $M$ that serves as a proxy for the real-world system dynamics. During learning, we wish to obtain a model $\hat{M}$ that learns the relative transition probabilities from a data set generated by observing the robot operate under the dynamics given by $M$. We assume sufficient domain knowledge is available to identify the existence of causal relationships between action, confounder, and state variables. Thus we assume the model structure is known and seek to learn the conditional probabilities. We further assume the transition function can be represented as a combination of a deterministic assignment function $f_{s^\prime}(s,\Delta s)$ of current state $s$ and relative state transition $\Delta s$, and a relative transition probability $P(\Delta S|A,U)$. We further assume the transition function can be partitioned into two components, one for when the agent is in a state $s\in S_{UC}$ where it is under the influence of confounding ($P_{UC}$) and the other where the agent is not ($P_{0}$), such that:
\begin{equation}
  P(S^\prime|S,A,U) =
  \begin{cases}
    P_{UC}(S^\prime|\Delta S,A,U) & \text{for $s\in S_{UC}$} \\
    P_{0}(S^\prime|\Delta S,A) & \text{otherwise,}
  \end{cases}
\end{equation}
and we learn the relative transition functions for each case: $P_{UC}(\Delta S|A,U)$ and $P_{0}(\Delta S|A)$. In the context of the \emph{GridWorldConfounded} problem, $P_{UC}$ describes the transition probabilities when the robot is located in the orange cell and thus affected by the electromagnet (Figure \ref{fig:grid_world_confounded}), while $P_{0}$ describes the transition probabilities when located in the other cells and thus not affected.

We further assume a two-phase data-collection approach, common in robot industrial inspection applications: an initial site survey with humans in the loop to collect observational robot data followed by autonomous robot deployment. In this setting we argue that it is reasonable to assume that the robot-and-human site survey team is equipped with sensors to observe all ground truth model variables in a privileged setting, including the confounder that is unobserved during autonomous deployment. Given this, we also seek to learn a parameterisation for $P(U)$, the independent distribution governing the probabilities of the confounding variable. We assume a sensor model is known, specifying $P(Z|S^\prime)$, and assume a known deterministic reward function specified by system designers. Thus, we do not learn these probabilities.

To begin the learning process, we sample a data set $D$ of $N$ number of observations from the ground truth model $M$, represented as a probabilistic program in the Python-based Pyro probabilistic programming language (PPL) \cite{Bingham2018Pyro}. Each observation is a record of $\langle {UC}, U, A, \Delta S\rangle$, where $UC$ is a Boolean value indicating whether the agent was under the influence of the confounder when selecting its choice of action. Next, we define a learning model $\hat{M}$ with the same causal structure as $M$, but with unknown parameterisation of $P(U)$, $P_{UC}(\Delta S|A,U)$, and $P_{0}(\Delta S|A)$ each represented as a categorical distribution. 
Finally, we learn the parameterisation of these three distributions by fitting $\hat{M}$ to the the data set $D$ through automated stochastic variational inference (SVI) \cite{Wingate2013VariationalInference, Ranganath2014BlackBoxVariationalInference} in Pyro PPL. We use the Adam optimizer \cite{Kingma2015AdamOptimizer} in PyTorch, and \emph{TraceEnum\_ELBO} which is a trace implementation of evidence lower bound (ELBO) based SVI in in Pyro.
We note that the confounder does not bias the learning method because the confounder is controlled for using the privileged observations of it.

\subsection{Causally-Informed POMDP Planning Method}
Here we describe CAR-DESPOT, a novel causally-informed extension of AR-DESPOT that combines the online planning performance of AR-DESPOT with the causal expressivity of SCMs to eliminate policy errors due to confounding bias. 

Since the SCM-UCPOMDPs subsume standard POMDP models, they retain properties of POMDPs and therefore can be used in their place. Thus, in CAR-DESPOT we formulate models as SCM-UCPOMDPs to take advantage of the existing POMDP-based search machinery in AR-DESPOT to find approximately optimal policies. 

However, we make a novel and fundamental change to how the transition function is sampled during exploration rollouts.
While AR-DESPOT samples from observational transition distribution $P(S^\prime|S, A)$, which is prone to errors from unobserved confounders, CAR-DESPOT samples from the interventional distribution $P\big(S^\prime|S,do(A=a)\big)$ to avoid bias errors. 
We anticipate this will improve the planner performance in two ways: 1) it will avoid the heuristic incorrectly guiding the incremental DESPOT construction towards exploring less promising areas of the tree; and, 2) it will prevent the planner, given a constructed DESPOT, from incorrectly valuing discovered action-observation outcomes, thus leading to sub-optimal policies.
The same interventional modification is applied to computing the POMDP observation function probability distribution. However, for models for which the observation is conditional only on the successor state, such as the toy model considered here,
$P(Z|S^\prime,A)$ and $P\big(Z|S^\prime,do(A=a)\big)$ are equivalent. 

We represent SCM-UCPOMDP models as causal generative probabilistic programs in Pyro PPL and use Importance Sampling inference \cite{Kloek1978ImportanceSampling} to compute interventional transition posterior distributions, used to inform planner rollouts.
We implement CAR-DESPOT by extending the C++ implementation of AR-DESPOT\footnote{AR-DESPOT - https://github.com/AdaCompNUS/despot} provided by authors in \cite{Somani2013DESPOT}. 

\section{EXPERIMENTS}
We evaluate our proposed model learning and causally-informed POMDP planning methods on a minimal toy problem inspired by the autonomous robot inspection domain.
\subsection{Problem Description}
\emph{GridWorldConfounded} (Fig. \ref{fig:grid_world_confounded}) is a modified version of the \emph{GridWorld} problem \cite{Sutton1990GridWorld} with an unobserved confounder, the presence of which is known to exist \textit{a priori}. The robot is located on a 2-dimensional grid and its aim is to navigate to a goal location to inspect an important asset as part of an industrial site inspection mission, while avoiding a collision with occupied cells.  The robot always begins at the bottom-left grid cell $(0,0)$ and knows the position of the goal which is always $(0,3)$. 
The robot can move one grid cell in each of the four cardinal directions.
Due to actuator errors, when executing its movement action there is a small chance the robot may drift and end up either to the left or right (with respect to the direction of travel) with probability of $5\%$ in either direction. 
The robot receives a noiseless observation of its position after each action.

Critically, an electromagnet is located in cell $(1,2)$ and affects the robot's orientation sensor measurements and action selection when in the magnet's region of influence, located to the left in $(0,2)$. The orientation error caused by the magnet $U_{OrientationError}$ fluctuates randomly over time and is not able to be perceived by the robot due to lack of a suitable sensor. 
At each time step, the orientation error is sampled from an independent categorical distribution over discrete values $\{-90\degree,0\degree,+90\degree\}$ with probabilities $[0.10, 0.80, 0.10]$. 

The unobserved confounder influences the robot in two ways.
First, when the robot is at $(0,2)$, the robot's reaction is to take an action with the probabilities given shown in Table \ref{tab:GridWorldConfoundedActionSelectionProbabilities}. 
These probabilities reflect the reflexive rules internal to the robot's reactive decision-making; here they are chosen to demonstrate the effects of confounding.
The second effect of the confounding variable is on the stochastic execution of the movement action. If under the influence of the magnet, the robot will attempt to move in its intended direction but because of orientation error, it will actually move with an offset given by the orientation influence. For example, when taking the \emph{RIGHT} action ($0\degree$ heading in grid frame) given a $+90\degree$ orientation error offset, the robot will instead move up. 
The drift dynamics are applied after the effects of confounding. When the robot is outside of the region of influence, only the drift transition probabilities apply.
\begin{table}[h]
	\centering
	\caption[Reactive action selection distribution for \emph{GridWorldConfounded}.]{The reactive action selection probability distribution for the \emph{GridWorldConfounded} problem, conditional on the unobserved confounding variable $U_{OrientationError}$.}
	\begin{tabular}{ p{0.41\columnwidth}  P{0.12\columnwidth}  P{0.12\columnwidth}  P{0.12\columnwidth}}
		\toprule
		$P(A|U_{OrientationError})$ & $-90\degree$ 	& $0\degree$ 	& $+90\degree$ 	\\ \midrule
		\emph{RIGHT}				& 0.05			& 0.45			& 0.05			\\ 
		\emph{UP}					& 0.85			& 0.05			& 0.85			\\ 
		\emph{LEFT}					& 0.05			& 0.45			& 0.05			\\ 
		\emph{DOWN}					& 0.05 			& 0.05			& 0.05			\\ \bottomrule
	\end{tabular}
	\label{tab:GridWorldConfoundedActionSelectionProbabilities}
\end{table}

Each movement action gives a base reward of $-1$. After an action, if the robot is at the goal it receives $+100$ reward and the scenario terminates, else if it is in collision it receives $-50$ reward and terminates; otherwise the problem continues.

While this is a simplified and highly abstract problem, the magnet acts as a demonstration of how unobserved confounders can feasibly arise in practice during robot deployments in complex and uncontrolled real-world  environments.

\subsection{Model Parameter Learning}
We generate a data set of size $N=800,000$ by sampling from the ground truth model $M$ and use this as an input to SVI training. We use a learning rate $lr=0.01$ and 5001 training steps
to obtain the learned model $\hat{M}$, including the parameterisation of: $P_{UC}(\Delta Position,A,U_{OrientationError})$, $P_{0}(\Delta Position,A)$, and $P(U_{OrientationError})$. We learn 106 latent model parameters, using uniform Dirichlet priors for categorical distribution parameters in our learning model and the \emph{AutoDelta} automatic guide for the SVI guide.
We report the final ELBO-loss, which based on the Kullback–Leibler (KL) divergence, and the KL divergence of the learned full transition probability distribution $\hat{P}(S^\prime|S,A,U)$.

\subsection{Planner Evaluation}
We evaluate CAR-DESPOT on the GridWorldConfounded toy problem and compare it with a baseline method OAR-DESPOT, a version of AR-DESPOT modified to be compatible with SCM-UCPOMDP model representation but using observational transition quantities as done in the original algorithm. To assess the impact of our proposed learning method on planning performance, we evaluate each method in 2 cases, in which: 1) the learned model $\hat{M}$ is used for planning but the ground truth model $M$ is used for policy execution as a proxy to the real-world; and, 2) the ground truth model is used for both planning and policy execution.

The performance is assessed according to the average total discounted reward 
over 100 independent online planner executions.
We define the maximum scenario length, maximum search depth, and policy simulation depth as 15 steps. A search budget of 150 seconds is given to the planners for each time step. This was chosen empirically to allow the planners to sample a sufficient number of trials during the incremental DESPOT creation to sample the low-probability transition outcomes leading to failure in the toy problem.
The default discount factor $\gamma=0.95$ and target gap ratio $\epsilon=0.95$ is used, with regularisation pruning constant $\alpha=0.01$. The number of scenarios is 500, a sufficient amount to achieve performance in similar problems in \cite{Ye2017DESPOT}. 
To avoid penalising the first step of online planning, all required transition inference queries are computed and memoized prior to planning. Observation function inference queries are computed as-needed during belief updates but cached for future re-use. 5000 particles are used in Importance Sampling for both inference types; this was chosen empirically to accurately infer probabilities for low-probability outcomes. All planner evaluations are performed on a desktop PC with Ubuntu 20.04 operating system, using CPU-only, on an AMD Ryzen 9 5900x 12-core processor with 128 GB of RAM.

\section{RESULTS AND DISCUSSION}
\subsection{Analysis of learned model}
We begin by presenting and analysing the partially-learned SCM-UCPOMDP model of the GridWorldConfounded problem. SVI training yields an ELBO-loss of 3.300M (4.125 per data point) and produces interventional relative distribution probabilities each with an absolute error of less than 0.01 from ground truth values. Further, the probabilities for $U_{OrientationError}$, $[0.10, 0.80, 0.10]$, are accurately recovered during training: $[0.1004, 0.7997, 0.0999]$. The KL divergence of the learned full transition probability distribution $\hat{P}(S^\prime|S,A,U)$ from ground truth distribution ${P}(S^\prime|S,A,U)$ is $D_{KL}=0.0021$, indicating a very close fit.

\begin{figure}[htp]
    \centering
    \includegraphics[trim=30 60 25 60, clip, width=1.0\columnwidth]{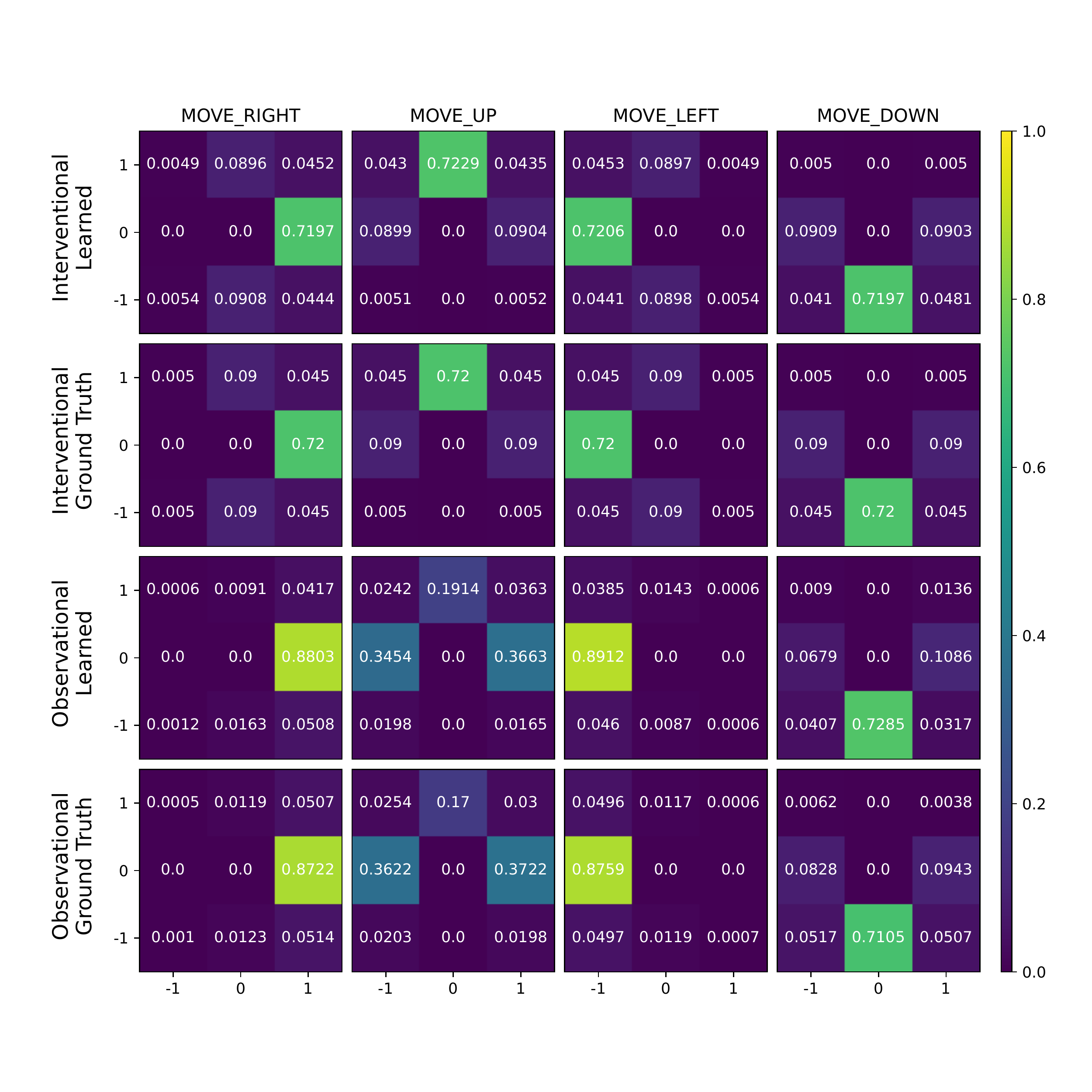}

    \caption{Interventional and observational transition probabilities of ground truth and learned models, under the influence of confounding, as a function of action choice. Formulating actions as causal interventions removes confounding bias errors present in in the observational transition predictions.}
    \label{fig:probability_heatmaps}
\end{figure}

Fig \ref{fig:probability_heatmaps} shows the learned parameters produce calculated \textbf{interventional} transition probabilities and inferred \textbf{observational} relative transition probabilities that are extremely close to the ground truth probabilities, when under the influence of confounding. Observational transition probabilities are inferred using Importance Sampling. 
These relative transition probabilities are provided as exemplar inference results; the full (i.e., non-relative) transition inference results used for planning are computed immediately prior to plan time.

All non-zero probability outcomes are captured by the learned model, even very low probability events (e.g., $p<0.001$), demonstrating that both the drift and magnet interference transitions dynamics are recovered accurately by the learning method.
The non-confounded observation and interventional transition probabilities are also recovered accurately. Due to space constraints these results are not shown as they are a subset of the confounded transition dynamics.

Critically, the learned model is able to capture the difference between the observational and interventional transition dynamics when under confounding. Comparing the probabilities shown in Fig \ref{fig:probability_heatmaps}, the learned model correctly captures the probability mass shift due to confounding bias in the observational transition dynamics. 
This is most evident for the up action, for which the probability mass of outcome $(0,+1)$ is redistributed to outcomes $(-1,0)$ and $(+1,0)$; this causes the observational distribution to significantly underestimate the interventional probability: 0.1914 instead of 0.7229. 
We expect that when the learned model is used for planning with OAR-DESPOT, the up action will be undervalued when under the effects of confounding. We expect this when using the ground truth model also.

\subsection{Analysis of planning performance}
\begin{table}
	\centering
    \caption[Algorithm Mean Performance Comparison.]{Mean total discounted rewards per model and algorithm combination. Standard errors are also given.}
	\begin{tabular}{ p{0.25\columnwidth}  P{0.30\columnwidth}  P{0.30\columnwidth}}
		\toprule
		{} 					& Learned Model			& Ground Truth Model			\\ \midrule
		OAR-DESPOT			& $ \ 0.74	 \pm  7.73 $			& $ 10.81	\pm  8.53  $	    \\ \midrule
		CAR-DESPOT			& $ 21.93 \pm  9.21 $	& $   15.54	 \pm  9.34$	\\  \bottomrule 
        %
	\end{tabular}
	\label{tab:mean_planning_performances}
    \vspace{-10pt}  
\end{table}

\begin{figure}[tp]
    \centering
    \begin{subfigure}{\columnwidth}
    \includegraphics[trim=10 0 0 0, clip, width=0.95\columnwidth]{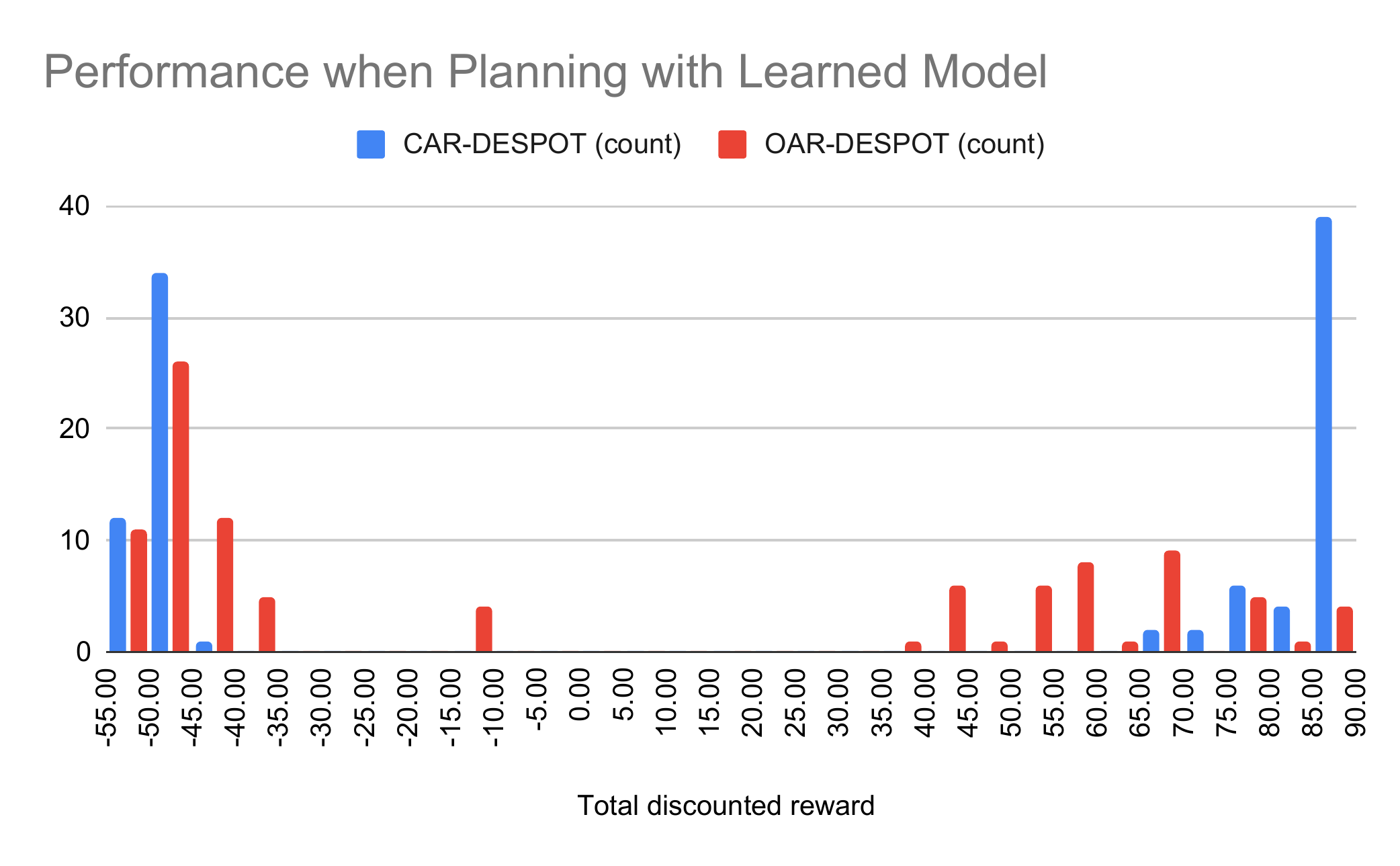}
    \end{subfigure}
    
    
    \begin{subfigure}{\columnwidth}
    \includegraphics[trim=10 0 0 0, clip, width=0.95\columnwidth]{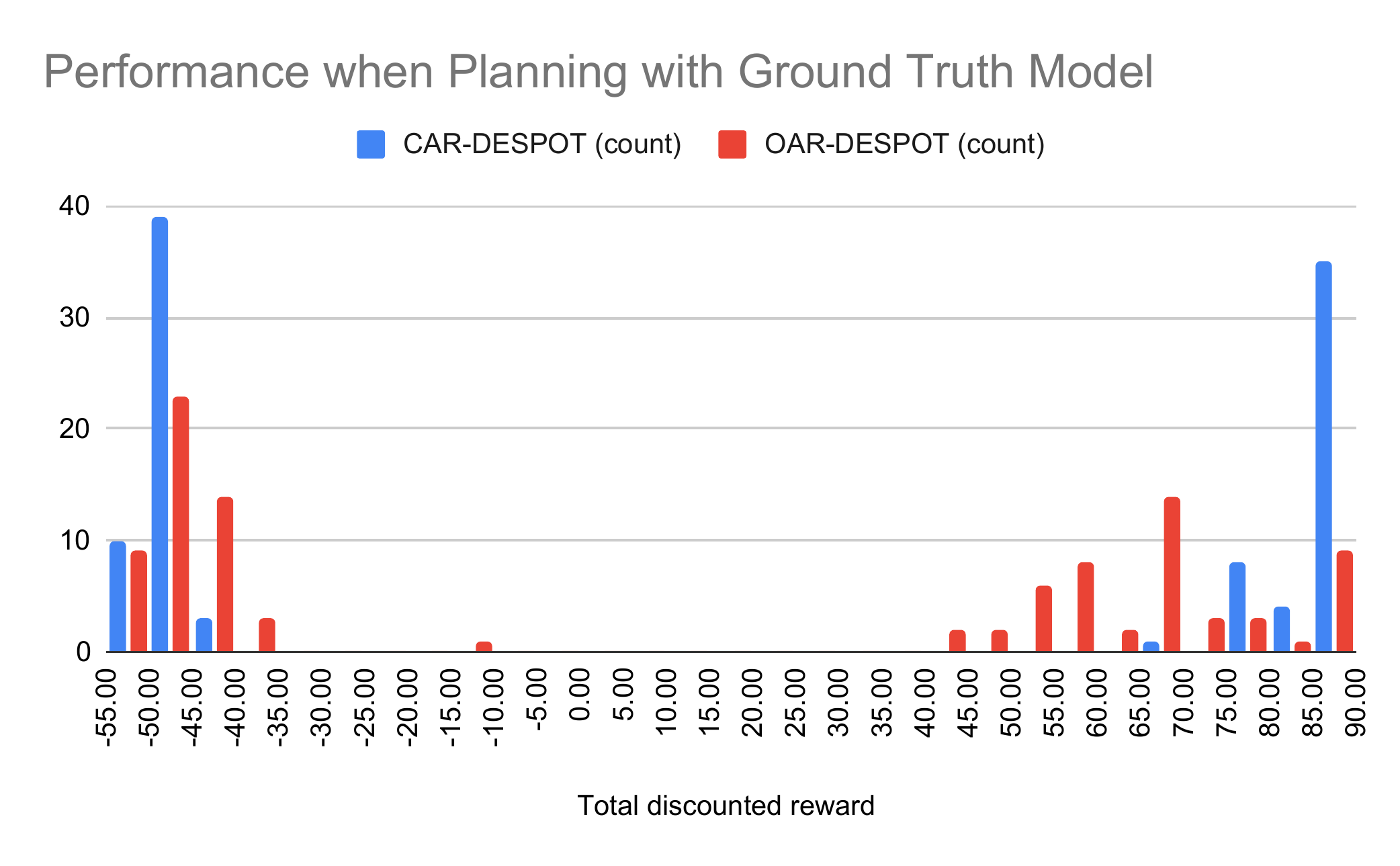}
    \end{subfigure}
    
    \caption{Distribution of CAR-DESPOT and baseline planner performance, with learned and ground truth models, according to total discounted reward. Policies from CAR-DESPOT perform better overall due to eliminating confounding bias.}
    \label{fig:planning_performance}
    \vspace{-15pt}  
\end{figure}

We present and discuss the planner evaluation results for the GridWorldConfounded problem. The distribution of policy performance over 100 runs of each planner and model combination is given in Fig \ref{fig:planning_performance}. We report the total discounted reward of policies executed on the ground truth model. These performances are summarised in Table \ref{tab:mean_planning_performances}.

The performance distributions of all planner-model combinations are largely bimodal, with rewards either clustered around the lower end of the reward range or upper end, corresponding to task failure by colliding with a wall or task success by reaching the goal respectively. Notably, for OAR-DESPOT there is also a small third peak around $-10$ corresponding to task failure by means of exhausting the maximum allowable number of steps. This occurred in 4 runs for learned model and 1 run for ground truth, in which the OAR-DESPOT-generated policies caused oscillatory behaviour that prevented reaching the goal within the action budget. This did not occur in any runs for CAR-DESPOT.

For the \textbf{learned model}, the task-success peak is more concentrated for CAR-DESPOT than OAR-DESPOT and has a higher mode. The reason for this is that since there is no confounding bias in the inferred interventional transition dynamics when the robot is at $(0,2)$ and thus under the influence of the confounder, CAR-DESPOT correctly estimates the value of taking actions to move along the left-hand side of the grid, which is shorter and therefore results in a higher score due to fewer movement penalties and cumulative rewards being discounted over a shorter time horizon. Thus, in all 100 runs CAR-DESPOT chooses to take the left-hand path. There is also another benefit of taking the shorter path: the cumulative probability of accidentally drifting into a wall resulting in task failure is lower due to the fewer steps taken. Thus, more runs achieve task success: 53 for CAR-DESPOT and 42 for OAR-DESPOT; this further increases the mean performance. Additionally, 39 runs achieve the maximum score. The lower tail end of the task-success peak reflects a small number of runs which exhibited minor oscillatory behaviour before reaching the goal. The task-failure peak is again more concentrated but with a lower mode. These reflect runs in which the outcomes of the correctly chosen actions by chance resulted in unintended outcomes, due either to the drift dynamics or the confounding variable influencing the outcome, leading to task failure. Since the CAR-DESPOT policies choose the left-hand path, failures occur earlier in the run and thus are discounted less.

Conversely, OAR-DESPOT underestimates the value of taking the left-hand path due to the confounding bias in the inferred transition probabilities for taking the up action at $(0,2)$ when under the influence of the confounder. 
Because of this, it has a strong preference to take the right-hand path to the goal, which is free from confounding but has a longer action cost and higher cumulative chance of failure due to drift. This yields a lower total discounted score. Interestingly, in 4 of the runs OAR-DESPOT takes the left-hand path and achieves the maximum possible score. 
This is likely due the incremental DESPOT construction missing either: 1) the failure outcomes at the confounder influence location; or, 2) the success outcome for taking the right-hand path.
%
This demonstrates that CAR-DESPOT is able to generate overall better performing policies than OAR-DESPOT.

Similar results are observed for the \textbf{ground truth model}. The CAR-DESPOT task-success peak is again narrower and concentrated around a higher value than that of OAR-DESPOT, while the task-failure peak is narrower and concentrated around a lower value, for the same reasons identified previously.
The number of task-success runs using the ground truth model with CAR-DESPOT is slightly lower compared to the learned model case: 48 compared to 53.
This slight difference in frequency-based counts is expected due to the stochastic nature of the problem \textemdash\ even executing an optimal policy can occasionally and randomly result in task failure. However, we expect that these task-success counts would converge in the limit of performing more evaluation runs.
Notwithstanding this, these results demonstrate that, when using the ground truth model, planning with CAR-DESPOT produces a better overall performance than OAR-DESPOT. Further, the similar results obtained from use of learned and ground truth models for planning further demonstrates the ability of our learning method to learn the relative transition dynamics accurately.

\subsection{Limitations and Improvements}
One limitation of our model learning method is the assumption that transition dynamics can be decomposed into a relative transition function and a known assignment function of the current state and relative state change. While this is suitable for the robot 2D position state representation in our toy problem, it may not be suitable for other state representations. In these cases, the full transition function of size $|S|\times|A|\times|S|$ will need to be learned, which will require a larger data set and an increased number of training steps.

We have evaluated our proposed method on a simple toy POMDP model with a relatively small state size (12), action space (4), and observation space (30). Despite this, the search time required to compute a good-performing policy online at each step is found to be 150s, which is prohibitively high for a mobile robot operating in real-time. 

Finally, analysis in \cite{Zhang2016CausalMDPUC} has shown that for MDP-based problems with unobserved confounders, optimising policies based on the interventional transition function is not always guaranteed to produce optimal policies. Rather, policy optimisation using counterfactual inference produces a dominant policy.
In this approach, the agent's reactive action selection is taken into account, which provides information about the state of the hidden confounder.
We expect that this improvement will translate to our work in POMDP planning.

\section{CONCLUSION}
In this paper we presented a novel causally-informed extension of the AR-DESPOT online POMDP planner, using causal modelling and inference to eliminate confounding bias from unmeasured confounding variables. Through experimental validation on a toy POMDP problem with unobserved confounding, we demonstrated that our method outperforms the baseline method. Using SCMs, we integrated causal reasoning into the POMDP formulation to enable robust POMDP-based planning for robots operating in real-world confounded environments.









\bibliographystyle{ieeetr}
\bibliography{references}

\begin{thebibliography}{10}

\bibitem{Kurniawati2022POMDPReview}
H.~Kurniawati, ``Partially observable markov decision processes and robotics,''
  {\em Annual Review of Control, Robotics, and Autonomous Systems}, vol.~5,
  no.~1, pp.~253--277, 2022.

\bibitem{Smallwood1973}
R.~D. Smallwood and E.~J. Sondik, ``The optimal control of partially observable
  markov processes over a finite horizon,'' {\em Operations research}, vol.~21,
  pp.~1071--1088, 1973.

\bibitem{Bai2015DESPOTAutonomousDriving}
H.~Bai, S.~Cai, N.~Ye, D.~Hsu, and W.~S. Lee, ``Intention-aware online pomdp
  planning for autonomous driving in a crowd,'' in {\em 2015 IEEE International
  Conference on Robotics and Automation (ICRA)}, pp.~454--460, 2015.

\bibitem{Budd2022}
M.~Budd, P.~Duckworth, N.~Hawes, and B.~Lacerda, ``Bayesian reinforcement
  learning for single-episode missions in partially unknown environments,'' in
  {\em 6th Conference on Robot Learning (CoRL 2022)}, OpenReview, 2022.

\bibitem{Goldhorn2018SearchAndTrack}
A.~Goldhoorn, A.~Garrell, R.~Alquézar, and A.~Sanfeliu, ``Searching and
  tracking people with cooperative mobile robots,'' {\em Autonomous robots},
  vol.~42, pp.~739--759, 2018.

\bibitem{Pearl2009Causality}
J.~Pearl, {\em {Causality: Models, reasoning, and inference, second edition}}.
\newblock Cambridge university press, 2009.

\bibitem{Bareinboim2015ConfoundedBandits}
E.~Bareinboim, A.~Forney, and J.~Pearl, ``Bandits with unobserved confounders:
  A causal approach,'' in {\em Advances in Neural Information Processing
  Systems}, pp.~1342--1350, 2015.

\bibitem{Zhang2016CausalMDPUC}
J.~Zhang and E.~Bareinboim, ``Markov decision processes with unobserved
  confounders: A causal approach,'' tech. rep., Technical Report R-23, Purdue
  AI Lab, 2016.

\bibitem{Ye2017DESPOT}
N.~Ye, A.~Somani, D.~Hsu, and W.~S. Lee, ``Despot: Online pomdp planning with
  regularization,'' {\em Journal of Artificial Intelligence Research}, vol.~58,
  pp.~231--266, 2017.

\bibitem{Somani2013DESPOT}
A.~Somani, N.~Ye, D.~Hsu, and W.~S. Lee, ``Despot: Online pomdp planning with
  regularization,'' in {\em Advances in Neural Information Processing Systems},
  vol.~26, 2013.

\bibitem{Ganguly2023}
N.~Ganguly, D.~Fazlija, M.~Badar, M.~Fisichella, S.~Sikdar, J.~Schrader,
  J.~Wallat, K.~Rudra, M.~Koubarakis, G.~K. Patro, W.~Z.~E. Amri, and W.~Nejdl,
  ``A review of the role of causality in developing trustworthy ai systems,''
  2023.

\bibitem{Hellstrom2021CausationInRobotics}
T.~Hellström, ``The relevance of causation in robotics: A review,
  categorization, and analysis,'' {\em Paladyn, Journal of Behavioral
  Robotics}, vol.~12, pp.~238--255, 2021.

\bibitem{Lake2017BuildingMachines}
B.~M. Lake, T.~D. Ullman, J.~B. Tenenbaum, and S.~J. Gershman, ``Building
  machines that learn and think like people,'' {\em Behavioral and Brain
  Sciences}, vol.~40, p.~e253, 11 2017.

\bibitem{Pearl2019CausalInferenceAndML}
J.~Pearl, ``The seven tools of causal inference, with reflections on machine
  learning,'' {\em Commun. ACM}, vol.~62, pp.~54--60, 2 2019.

\bibitem{Uhde2020RobotAsScientist}
C.~Uhde, N.~Berberich, K.~Ramirez-Amaro, and G.~Cheng, ``{The robot as
  scientist: Using mental simulation to test causal hypotheses extracted from
  human activities in virtual reality},'' in {\em IEEE International Conference
  on Intelligent Robots and Systems}, pp.~8081--8086, IEEE, 2020.

\bibitem{Diehl2022WhyDidIFail}
M.~Diehl and K.~Ramirez-Amaro, ``Why did i fail? a causal-based method to find
  explanations for robot failures,'' {\em IEEE Robotics and Automation
  Letters}, vol.~7, pp.~8925--8932, 2022.

\bibitem{Beetz1998Causal}
M.~Beetz and H.~Grosskreutz, ``{Causal Models of Mobile Service Robot
  Behavior},'' in {\em Proceedings of AIPS}, pp.~163--170, Link{\"{o}}ping
  University Electronic Press, 1998.

\bibitem{Aker2011Causal}
E.~Aker, A.~Erdogan, E.~Erdem, and V.~Patoglu, ``{Causal reasoning for planning
  and coordination of multiple housekeeping robots},'' in {\em Lecture Notes in
  Computer Science (including subseries Lecture Notes in Artificial
  Intelligence and Lecture Notes in Bioinformatics)}, vol.~6645 LNAI,
  pp.~311--316, Springer, 2011.

\bibitem{Diehl2023ExplainPredictPreventFailures}
M.~Diehl and K.~Ramirez-Amaro, ``A causal-based approach to explain, predict
  and prevent failures in robotic tasks,'' {\em Robotics and Autonomous
  Systems}, vol.~162, p.~104376, 2023.

\bibitem{Chen2016POMDPLite}
M.~Chen, E.~Frazzoli, D.~Hsu, and W.~S. Lee, ``Pomdp-lite for robust robot
  planning under uncertainty,'' in {\em 2016 IEEE International Conference on
  Robotics and Automation (ICRA)}, pp.~5427--5433, 5 2016.

\bibitem{Kurniawati2016}
H.~Kurniawati and V.~Yadav, {\em An Online POMDP Solver for Uncertainty
  Planning in Dynamic Environment}, pp.~611--629.
\newblock Cham: Springer International Publishing, 2016.

\bibitem{Silver2010POMCP}
D.~Silver and J.~Veness, ``Monte-carlo planning in large pomdps,'' {\em
  Advances in neural information processing systems}, vol.~23, 2010.

\bibitem{Ortega2021}
P.~A. Ortega, M.~Kunesch, G.~Delétang, T.~Genewein, J.~Grau-Moya, J.~Veness,
  J.~Buchli, J.~Degrave, B.~Piot, J.~Pérolat, T.~Everitt, C.~Tallec,
  E.~Parisotto, T.~Erez, Y.~Chen, S.~E. Reed, M.~Hutter, N.~de~Freitas, and
  S.~Legg, ``Shaking the foundations: delusions in sequence models for
  interaction and control,'' {\em CoRR}, vol.~abs/2110.10819, 2021.

\bibitem{Gasse2021}
M.~Gasse, D.~Grasset, G.~Gaudron, and P.-Y. Oudeyer, ``Causal reinforcement
  learning using observational and interventional data,'' 2021.

\bibitem{Pearl2018Book}
J.~Pearl and D.~Mackenzie, {\em {The book of why: the new science of cause and
  effect}}.
\newblock Basic books, 2018.

\bibitem{Madani1999Undecidability}
O.~Madani, S.~Hanks, and A.~Condon, ``On the undecidability of probabilistic
  planning and infinite-horizon partially observable markov decision
  problems,'' in {\em AAAI/IAAI}, pp.~541--548, 1999.

\bibitem{Bingham2018Pyro}
E.~Bingham, J.~P. Chen, M.~Jankowiak, F.~Obermeyer, N.~Pradhan, T.~Karaletsos,
  R.~Singh, P.~Szerlip, P.~Horsfall, and N.~D. Goodman, ``{Pyro: Deep universal
  probabilistic programming},'' {\em Journal of Machine Learning Research},
  vol.~20, 2019.

\bibitem{Wingate2013VariationalInference}
D.~Wingate and T.~Weber, ``Automated variational inference in probabilistic
  programming,'' 1 2013.

\bibitem{Ranganath2014BlackBoxVariationalInference}
R.~Ranganath, S.~Gerrish, and D.~M. Blei, ``Black box variational inference,''
  {\em Journal of machine learning research}, vol.~33, pp.~814--822, 2014.

\bibitem{Kingma2015AdamOptimizer}
D.~P. Kingma and J.~Ba, ``Adam: A method for stochastic optimization,'' in {\em
  3rd International Conference for Learning Representations}, 2015.

\bibitem{Kloek1978ImportanceSampling}
T.~Kloek and H.~K. van Dijk, ``Bayesian estimates of equation system
  parameters: An application of integration by monte carlo,'' {\em
  Econometrica}, vol.~46, no.~1, pp.~1--19, 1978.

\bibitem{Sutton1990GridWorld}
R.~S. Sutton, ``Integrated architectures for learning, planning, and reacting
  based on approximating dynamic programming,'' in {\em Proceedings of the 7th
  International Conference on Machine Learning (1990)}, pp.~216--224, Elsevier,
  1990.

\end{thebibliography}

\end{document}